\documentclass[letterpaper,conference]{ieeeconf}

\IEEEoverridecommandlockouts

\usepackage{amsmath}
\usepackage{amsfonts}
\usepackage{mathrsfs}
\usepackage{algorithm}
\usepackage{algorithmic}

\usepackage{mwe}
\usepackage{array}
\usepackage{graphicx}
\usepackage{booktabs}
\usepackage{multirow}
\usepackage{makecell}
\usepackage[flushleft]{threeparttable}
\usepackage{tablefootnote}
\usepackage{stfloats}
\usepackage{float}

\usepackage[
    caption=false,
    font=normalsize,
    labelfont=sf,
    textfont=sf
]{subfig}

\usepackage{textcomp}
\usepackage{url}
\usepackage{verbatim}
\usepackage{xcolor}   % <-- 用 xcolor，不要只用 color
\usepackage{cite}
\usepackage{balance}
\usepackage{xpatch}
\usepackage{etoolbox}

\usepackage{tikz}
\usetikzlibrary{calc,arrows,decorations.markings}

\usepackage[
    colorlinks=true,
    linkcolor=blue,
    citecolor=blue,
    urlcolor=blue
]{hyperref}

\newcommand{\best}[1]{\textbf{#1}}
\newcommand{\ours}{\textsc{$\boldsymbol{\varepsilon}\mathrm{4P}$}}

\graphicspath{{./imgs/}}

\makeatletter

\patchcmd{\@makecaption}
  {\scshape}
  {}
  {}
  {}

\patchcmd{\@makecaption}
  {\\}
  {.\ }
  {}
  {}

\makeatother

\title{\LARGE \bf
Imperfection for Precision: \\Upcycling Imperfect Data for High-Precision Robotic Manipulation}

\author{
    Hao Wei\textsuperscript{1*},
    Yang Liu\textsuperscript{1*},
    Chao Tang\textsuperscript{1*},
    Shengbao Li\textsuperscript{1},
    Jiangtao Chen\textsuperscript{1},
    Jinxuan Zhu\textsuperscript{1,2},
    Jiaheng Wang\textsuperscript{1}, \\
    Hong Yin\textsuperscript{1},
    Zhaofeng Cao\textsuperscript{1,3},
    and Tingguang Li\textsuperscript{1$\dagger$} \\[0.5em]
    \small
  \textsuperscript{1}Samsung Robotics eXperience
  \quad
  \textsuperscript{2}National University of Singapore
  \quad
 \textsuperscript{3}University of Illinois Urbana-Champaign
  }

\begin{document}

\maketitle
\thispagestyle{empty}
\pagestyle{empty}

\begingroup
\renewcommand{\thefootnote}{}
\footnotetext{\textsuperscript{*}Equal contribution.\hspace{3em}\textsuperscript{$\dagger$}Corresponding author.}
\endgroup

\maketitle

\thispagestyle{empty}
\pagestyle{empty}

\begin{abstract}

Training vision-language-action (VLA) models for high-precision manipulation typically requires task-specific, high-quality data (e.g., teleoperation), which is slow and expensive to collect. To reduce this burden without compromising manipulation precision, we propose $\boldsymbol{\varepsilon}\mathrm{4P}$ (Imperfection for Precision), a simple yet effective method that ``upcycles'' two otherwise discarded data sources: (1) low-precision data from the target task and (2) high-precision data from mismatched tasks. Rather than naively mixing these imperfect data sources throughout co-training, $\boldsymbol{\varepsilon}\mathrm{4P}$ controls where each source contributes along the flow-matching trajectory. Specifically, low-precision, target-task data is used at high noise to preserve high-level task context and high-precision, task-mismatched data is used at low noise to transfer low-level action precision. Through real-robot experiments on both sub-millimeter, high-precision tasks and coarse-grained tasks, we demonstrate that the proposed method (1) effectively leverages additional imperfect data to improve policy performance by up to 31.7 percentage points, and (2) can replace an equal amount of task-specific, high-quality data with an average performance drop of only 4.2 percentage points. Overall, $\boldsymbol{\varepsilon}\mathrm{4P}$ points toward a scalable paradigm for high-precision manipulation, in which heterogeneous, imperfect data can be systematically repurposed to reduce reliance on costly task-specific, high-quality data. More details are available at
\nolinkurl{https://varepsilon4p.github.io/}.

\end{abstract}

\section{Introduction}\label{sec:intro}

% 这是一些在abstract放不下的句子，可考虑从放到intro里
% In this work, we ask: How can we leverage imperfect data to reduce this burden without compromising manipulation precision? 

% admits UMI data at high noise, preserving task intent while suppressing low-level imprecision, and cross-task data at low noise, enabling precise motion transfer while minimizing interference from mismatched intent. 

% Through real-robot experiments on three high-precision manipulation tasks: bolt-and-nut sorting, cable plugging, and ATX 24-pin connector insertion, we demonstrate that $\boldsymbol{\varepsilon}\mathrm{4P}$ (1) 

% 1) high-quality teleoperated demonstrations from mismatched tasks, which provide precise low-level motion $\boldsymbol{\varepsilon}\mathrm{4P}$itives, and (2) low-precision UMI demonstrations of the target task, which provide aligned high-level task intent.

% 1. 现状，指出问题的重要性 (TC: 这段修改完毕)
Vision-language-action (VLA) models offer a promising path toward general-purpose robot manipulation by scaling on diverse robot experience~\cite{openvla,pi0,pi05}. However, training these models for high-precision manipulation tasks, such as insertion and assembly~\cite{zhu2025shapeforce}, still requires task-specific, high-quality data (i.e., teleoperation), as the tight geometric tolerances of these tasks demand highly accurate and consistent demonstrations. Collecting such data requires substantial human effort and hardware resources~\cite{droid,openx,tang2025mimicfunc,li2025demomodality,umi}. Meanwhile, large amounts of imperfect data may already be available, yet their quality often falls short of the stringent requirements of high-precision tasks, making them difficult to reuse directly~\cite{dataquality,mao2026beyond}. We therefore ask: \textbf{How can imperfect data be effectively upcycled for high-precision manipulation?}

% 2. 简单综述现有方法（引用），diss别人方法的不好 (TC: 这段修改完毕)
Existing methods primarily address this problem through native co-training, using sampling ratios or mixture weights to control how much each data source contributes while applying its contribution uniformly across the training objective~\cite{openx,pi05,remix,wu2019imperfect,wang2021weight,dwbc}. While effective for less demanding tasks, such uniform integration of imperfect data can introduce conflicting supervision and bias the policy away from the precise behaviors, a limitation also observed in prior work~\cite{adp}. Collecting substantially more task-specific, high-quality data could in principle mitigate this issue, but doing so is costly and difficult to scale. Conversely, simply increasing the amount of imperfect data does not necessarily resolve the mismatch and may even amplify conflicting supervision. This creates a practical dilemma: the most reliable data are difficult to scale, while the most scalable data cannot be naively integrated.

% \begin{figure}[t]
%   \centering
%   \includegraphics[width=\linewidth]{imgs/teaser/teaser_chao.png}
%   \caption{\textbf{Overview of $\boldsymbol{\varepsilon}\mathrm{4P}$ }: }
%   \label{fig:teaser}
% \end{figure}

\begin{figure}[t]
  \centering
  % \vspace*{0.1in}
  \begin{tikzpicture}[inner sep = 0pt, outer sep = 0pt]
    \node[anchor=south west] (fnC) at (0in,0in)
      {\includegraphics[height=2.7in,clip=true,trim=0.4in 0in 0in 0in]{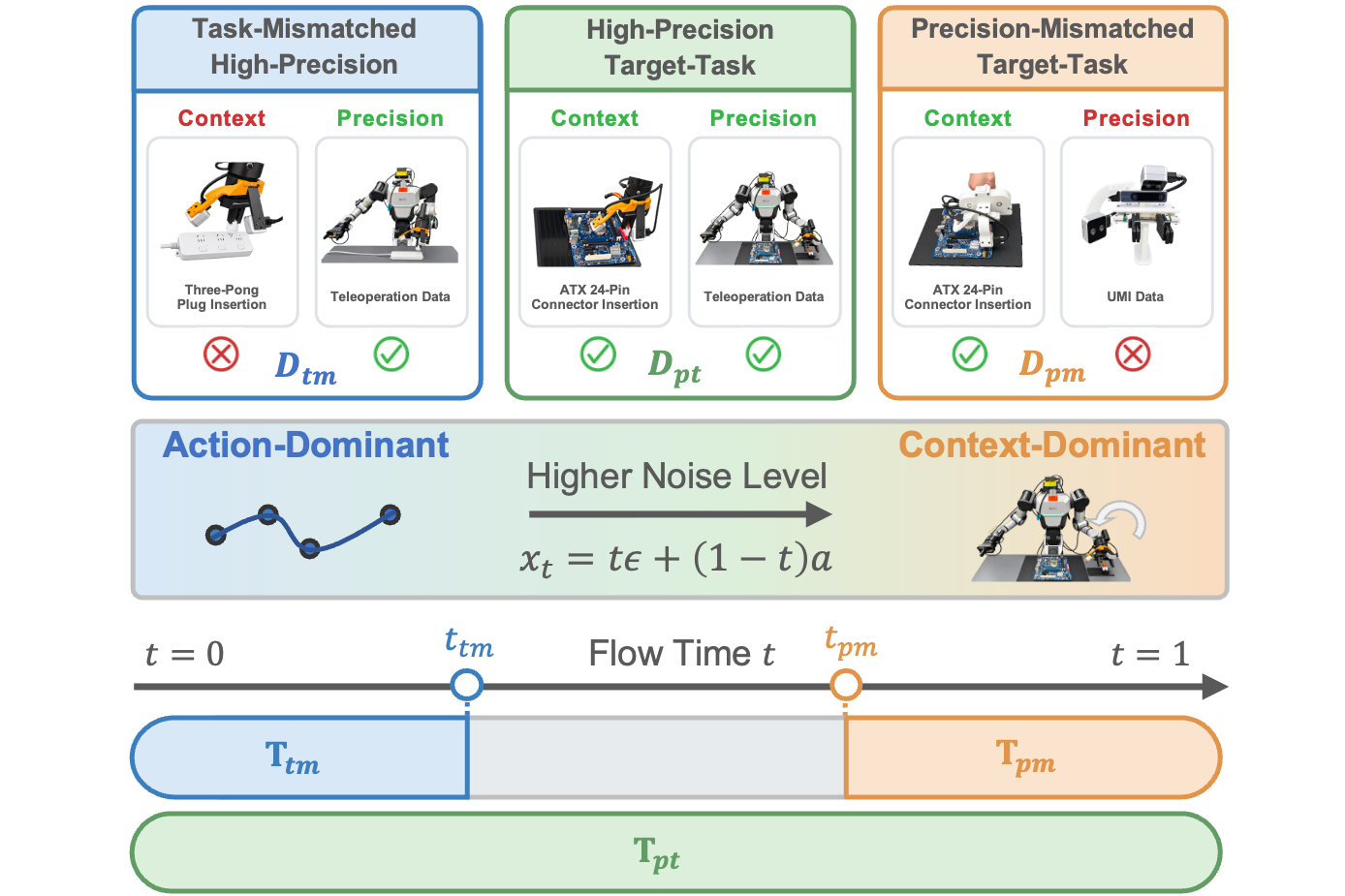}};
  \end{tikzpicture}
    \vspace*{-0.2in}
        \caption{\textbf{Conceptual overview of $\boldsymbol{\varepsilon}\mathrm{4P}$.} The proposed method upcycles two types of imperfect data sources by routing them to source-dependent flow time admission regions for high-precision manipulation.}
  \label{fig:teaser}
  % \vspace*{-0.3in}
\end{figure} 

 % The method routes high-precision but task-mismatched demonstrations $\mathcal{D}{\mathrm{tm}}$ to the action-dominant, low-noise region $\mathcal{T}{\mathrm{tm}}=[0,t_{\mathrm{tm}}]$, and precision-mismatched but task-aligned demonstrations $\mathcal{D}{\mathrm{pm}}$ to the context-dominant, high-noise region $\mathcal{T}{\mathrm{pm}}=[t_{\mathrm{pm}},1]$; high-precision target-task demonstrations $\mathcal{D}_{\mathrm{pt}}$ supervise the full flow trajectory. The boundaries are selected offline from complementary precision- and task-mismatch diagnostics.

% 3. 介绍自己的方法
In this work, rather than treating imperfect data as ``trash'', we argue that the key is to determine where along the flow-matching trajectory each data source can provide useful supervision. Based on this insight, we propose $\boldsymbol{\varepsilon}\mathrm{4P}$ (Imperfection for Precision), which routes imperfect data to source-dependent admission regions along the flow-matching trajectory of a policy~\cite{flowmatching} (e.g., a VLA), as illustrated in Figure~\ref{fig:teaser}. We consider two complementary sources of imperfect data: (1) low-precision target-task data collected with UMI~\cite{umi} and (2) high-precision teleoperation data from mismatched tasks. To determine the admission region for each source, $\boldsymbol{\varepsilon}\mathrm{4P}$ uses two complementary mechanisms tailored to their respective imperfections. For the former, we identify the flow-time region where its noised actions become sufficiently indistinguishable from high-precision demonstrations of the same task. For the latter, we identify the transition from context-dominant prediction at high noise to action-dominant prediction at low noise along the flow-time axis. By restricting each source to an appropriate flow-time region, $\boldsymbol{\varepsilon}\mathrm{4P}$ effectively integrates heterogeneous imperfect data into a single policy while limiting their imperfections, without modifying the model architecture or flow-matching objective.

% 4. 做了哪些实验，效果如何
To demonstrate the effectiveness of $\boldsymbol{\varepsilon}\mathrm{4P}$, we conduct real-robot experiments on both sub-millimeter, high-precision tasks (two-stage cable plugging and ATX 24-pin connector insertion) and coarse-grained manipulation tasks (Bolt-Nut Sorting). The results show that $\boldsymbol{\varepsilon}\mathrm{4P}$ (1) effectively leverages additional imperfect data to improve policy performance by up to 31.7 percentage points, and (2) can replace an equal amount of task-specific, high-quality data with an average performance drop of only 4.2 percentage points. Moreover, $\boldsymbol{\varepsilon}\mathrm{4P}$ consistently outperforms native co-training and source reweighting strategies commonly used in prior work.

% 5. 列contribution，做实验不算contribution，只说技术创新
\textbf{Contribution.} We propose a novel method that upcycles heterogeneous imperfect data by routing them to source-dependent admission regions along the flow-matching trajectory, reducing reliance on costly task-specific, high-quality data and providing a scalable paradigm for high-precision manipulation. 

\section{Related Work}\label{related}

\begin{figure*}[t]
    \centering
    \vspace{-0.2in}
    \makebox[\linewidth][c]{%
        \includegraphics[width=0.95\linewidth]{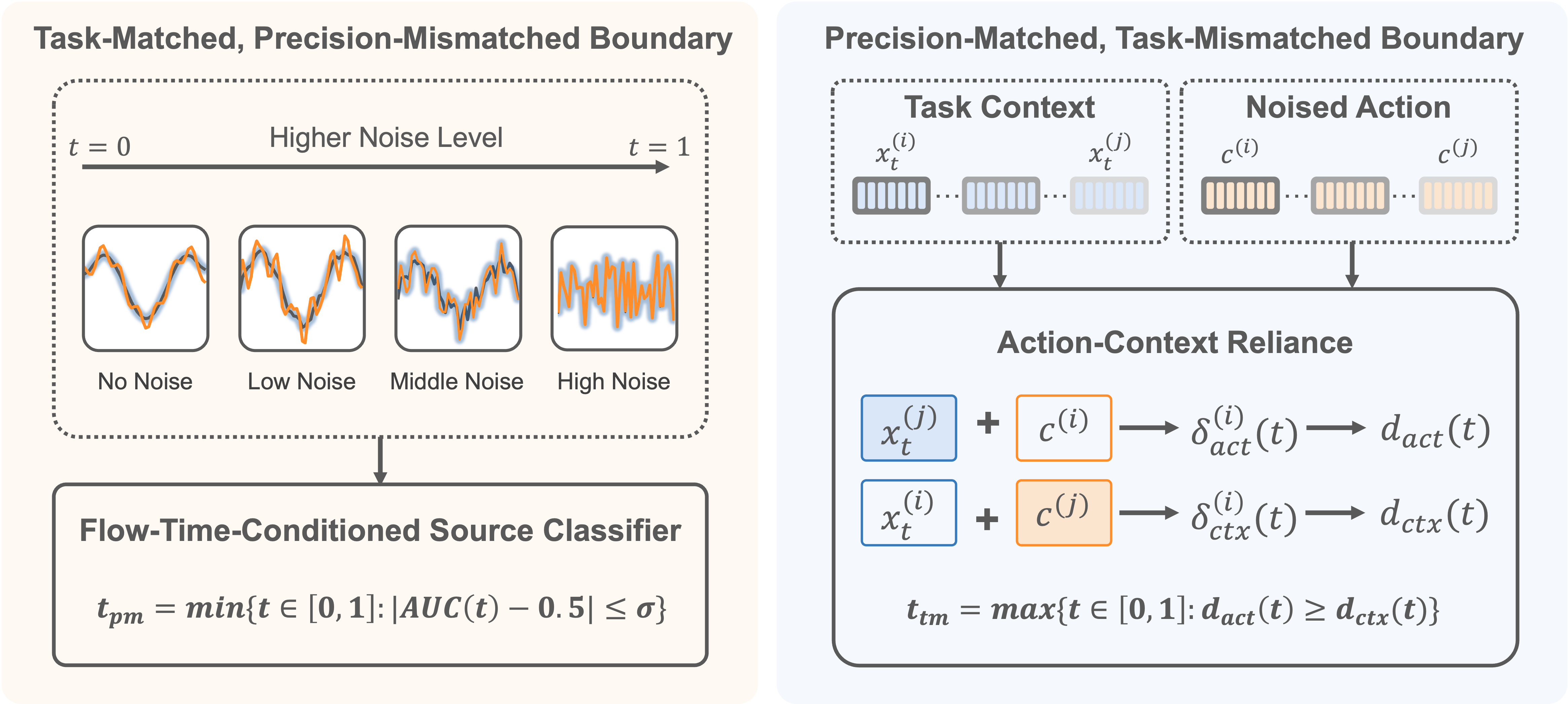}
    }
% \caption{Offline diagnostics for source-dependent flow-time admission regions. (a) A flow-time-conditioned source classifier compares noised actions from precision-mismatched and high-precision target-task data; $t_{\mathrm{pm}}$ is the earliest time at which the held-out AUC is indistinguishable from chance. (b) For task-mismatched, high-precision data, interventional swaps of context and noised action yield $d_{\mathrm{ctx}}(t)$ and $d_{\mathrm{act}}(t)$; $t_{\mathrm{tm}}$ is the latest time in the action-dominant regime, where $d_{\mathrm{act}}(t)\ge d_{\mathrm{ctx}}(t)$.
% }

\caption{An overview of two mechanisms for deriving the source-dependent admission boundaries $t_{\mathrm{pm}}$ and $t_{\mathrm{tm}}$, tailored to task-matched, precision-mismatched
  data $\mathcal{D}_{\mathrm{pm}}$ (left) and task-mismatched, precision-matched data $\mathcal{D}_{\mathrm{tm}}$ (right), respectively. }
    \label{fig:method}
\end{figure*}

% \begin{figure*}[t]
%     \centering
%     \makebox[\linewidth][c]{%
%         \includegraphics[width=0.8\linewidth]{imgs/method/method.png}
%     }
% \caption{Offline diagnostics for source-dependent flow-time admission regions. (a) A flow-time-conditioned source classifier compares noised actions from precision-mismatched and high-precision target-task data; $t_{\mathrm{pm}}$ is the earliest time at which the held-out AUC is indistinguishable from chance. (b) For task-mismatched, high-precision data, interventional swaps of context and noised action yield $d_{\mathrm{ctx}}(t)$ and $d_{\mathrm{act}}(t)$; $t_{\mathrm{tm}}$ is the latest time in the action-dominant regime, where $d_{\mathrm{act}}(t)\ge d_{\mathrm{ctx}}(t)$.
% }
%     \label{fig:method}
% \end{figure*}

% 1。 多种数据混训方法总结，最后一句话指出缺点，体现PRIM的优势

\subsection{Heterogeneous Data Co-Training}

Generalist robot policies increasingly leverage heterogeneous training data from diverse tasks, domains, and data sources~\cite{openx,octo,tri_cotraining,wang2024poco,wang2024hpt}. Existing methods improve such data mixtures mainly through (1) reweighting at the domain level~\cite{remix} and (2) filtering or selecting useful demonstrations at the sample level~\cite{robotdatacuration,cupid}. For co-training with imperfect data, prior work has further explored demonstration reweighting, data or trajectory-segment selection, and training under systematic domain shifts~\cite{wang2021weight,dwbc,ssdf,simrealcotraining,simrealanalysis,kuhar2023discern,yue2024diverse,chen2025s2i}.
 These approaches show that imperfect data can still provide useful supervision when its contribution is appropriately controlled. However, they primarily regulate how much each domain, trajectory, or sample contributes, while applying its contribution uniformly across the training objective. Such uniform integration of imperfect data can introduce conflicting supervision~\cite{gandhi2023compatible} and bias the policy away from the precise behaviors required for high-precision manipulation.
 In contrast, $\boldsymbol{\varepsilon}\mathrm{4P}$ determines  where along the flow-matching trajectory each imperfect data source should contribute, routing it to source-dependent admission regions where it can provide useful supervision.

% \subsection{Data Selection and Transfer in Robot Learning}

% Robot learning increasingly relies on selecting useful supervision from
% large and heterogeneous demonstration collections. Existing approaches
% improve data mixtures by learning domain-level sampling weights
% ~\cite{remix}, or by estimating the quality or downstream influence of
% individual demonstrations for filtering and selection
% ~\cite{robotdatacuration,cupid}. A related line of work directly learns
% from non-expert or partially compatible data, for example by re-weighting
% suboptimal demonstrations~\cite{dwbc}, retaining useful segments from
% imperfect trajectories~\cite{ssdf}, or co-training with imperfect data
% under systematic domain shift~\cite{simrealanalysis}. 

% Together, these
% methods demonstrate that data outside the target expert distribution can
% still provide useful supervision. However, these approaches primarily assess data utility at the domain,
% trajectory, or sample level. PRIM introduces a complementary granularity: the same imperfect source can contribute differently across flow time,
% allowing transferable information to be retained only where its
% source-specific defect is less influential.

% 2. Noise-Level Scheduling：首先说理论理论基础，以及在其他领域的应用，承认我们沿袭了这个思路；说在机器人上的应用，主要是ADP，以及三点我们和ADP的不一样：（1）出发点：ADP从数据侧分析频谱得到，我们则是通过policy对at和condition在不同噪声步的敏感度不同的得到, (2)  ADP主要通过sweep计算tmax，我们提除了自己的方法

\subsection{Noise and Flow-Time Scheduling}

Diffusion and flow-matching models have been widely applied to generative modeling, where noise level provides a natural control axis over the generation process~\cite{ddpm,flowmatching}. In diffusion-based image restoration, this structure has been exploited for super-resolution, deblurring, and inpainting~\cite{ddrm,diffpir}, while forward diffusion can progressively attenuate unknown degradations before recovering the underlying image content~\cite{dr2}. More broadly, noise-dependent objective weighting shows that different diffusion/flow-time regions need not contribute equally during training~\cite{p2,minsnr}. Related work further shows that the utility of lower-quality or biased
data can vary across diffusion times~\cite{ambientomni}.
Together, these works suggest that the value of information and the influence of corruption vary systematically along the diffusion or flow-matching trajectory. 

In robotics, diffusion models have been widely adopted in sequential generation and robot policy learning~\cite{diffusionpolicy}, and recent work has begun to exploit non-uniform noise schedules for action generation. Diffusion Forcing introduces per-token diffusion schedules for sequence modeling~\cite{diffusionforcing}, while Streaming Diffusion Policy applies variable noise levels across the action horizon~\cite{streamingdp}. TMRL further modulates diffusion timesteps to control conditioning strength
during policy adaptation~\cite{tmrl}. These approaches demonstrate that different elements of a sequence can benefit from different noise levels, extending noise-dependent control from image generation to sequential decision making. 
Concurrent work, Ambient Diffusion Policy (ADP)~\cite{adp},
similarly exploits diffusion time to selectively leverage suboptimal data.
Taking a step forward, $\boldsymbol{\varepsilon}\mathrm{4P}$ provides a complete mechanism to automatically derive source-dependent admission regions for precision and task mismatch.

\section{Approach}\label{sec:method}

\subsection{Problem Formulation}\label{sec:overview}

We consider training a flow-matching policy (e.g., a VLA) using high-precision target-task data $\mathcal{D}_{\mathrm{pt}}$ together with two forms of imperfect supervision: (1) precision-mismatched target-task data $\mathcal{D}_{\mathrm{pm}}$, and (2) task-mismatched high-precision data $\mathcal{D}_{\mathrm{tm}}$. In our setup, these two sources are instantiated using UMI and teleoperation data, respectively.

Let $s\in\{\mathrm{pt},\mathrm{pm},\mathrm{tm}\}$ denote a data source, $\pi$ the source sampling distribution, and $t\in[0,1]$ the flow time. We associate each source with an admission region $\mathcal{T}_s\subseteq[0,1]$. For $\mathcal{D}_{\mathrm{pt}}$, we fix $\mathcal{T}_{\mathrm{pt}}=[0,1]$. Given $s\sim\pi$, the flow time is sampled uniformly from the corresponding region, $t\sim\mathcal{U}(\mathcal{T}_s)$. We refer to the unmodified setting in which $\mathcal{T}_s=[0,1]$ for all sources as \emph{native co-training}. This setting retains the original $\pi$ and applies neither source-dependent admission nor loss reweighting. Our objective is to determine $\mathcal{T}_{\mathrm{pm}}$ and $\mathcal{T}_{\mathrm{tm}}$ such that each imperfect source contributes only where it provides useful supervision.

In what follows, we first examine how the policy’s reliance on the noised action input and task context varies across flow time, providing a lens into how precision and task mismatch are expressed along the flow trajectory. We then derive source-dependent criteria for $\mathcal{T}_{\mathrm{pm}}$ and $\mathcal{T}_{\mathrm{tm}}$.

\subsection{How Flow Time Shifts Action-Context Reliance}
\label{sec:input-reliance}

 % by measuring the policy's reliance on these two inputs across flow time.

$\boldsymbol{\varepsilon}\mathrm{4P}$ builds on the observation that the policy's reliance on the noised action input and task context varies across flow time. We characterize this behavior by conducting an empirical analysis on $\pi_{0.5}$~\cite{pi05}, while the underlying observation is expected to extend to other flow-matching policies. Let $a\in\mathbb{R}^{H\times d}$ denote an action chunk and $c$ the task context, including visual observations, language instruction, and proprioception. We adopt the convention that $t=0$ corresponds to the clean action and $t=1$ to pure noise. Given $\epsilon\sim\mathcal{N}(0,I)$, the flow path is:
\begin{equation}
x_t=t\epsilon+(1-t)a
\label{eq:ot-path}
\end{equation}
The predicted action is therefore denoted as:
\begin{equation}
\hat a(x_t,c,t)
=
x_t-t\hat v(x_t,c,t)
\label{eq:ahat}
\end{equation}
%
% $\hat a(x_t,c,t)$. 
where $\hat v$ is the predicted flow velocity. We use $\hat a$ as an action-space readout to measure the policy's reliance on the noised action input $x_t$ and task context $c$ across flow time, quantified by the change in $\hat a$ under input perturbations.

% Using the predicted flow velocity $\hat v(x_t,c,t)$, we obtain the predicted action as
% %
% \begin{equation}
% \hat a(x_t,c,t)
% =
% x_t-t\hat v(x_t,c,t)
% \label{eq:ahat}
% \end{equation}
% %

% a different target-task episode
% For each sample $i$, rather than applying synthetic perturbations that may push the policy input out of distribution, We draw a reference sample $j$ from $\mathcal{D}_{\mathrm{pt}}$ and use $x_t^{(j)}$ or $c^{(j)}$ as an in-distribution proxy for perturbing $x_t^{(i)}$ or $c^{(i)}$, respectively. 
Specifically, we measure this input reliance using a Two-Stage Cable Plugging dataset with 100 episodes. For each sample $i$, we draw a reference sample $j$ from the dataset and substitute either $x_t^{(i)}$ or $c^{(i)}$ with its counterpart from $j$. This keeps the substituted component marginally in-distribution while measuring the policy's sensitivity to each input. The resulting change in $\hat a$ can be measured as:
\begin{align}
\delta_{\mathrm{act}}^{(i)}(t)
&=
\left\|
\hat a(x_t^{(i)},c^{(i)},t)
-
\hat a(x_t^{(j)},c^{(i)},t)
\right\|_2^2 \\
\delta_{\mathrm{ctx}}^{(i)}(t)
&=
\left\|
\hat a(x_t^{(i)},c^{(i)},t)
-
\hat a(x_t^{(i)},c^{(j)},t)
\right\|_2^2
\end{align}
Here, $\delta_{\mathrm{act}}$ measures the sensitivity to the noised action input, whereas $\delta_{\mathrm{ctx}}$ measures the sensitivity to task context. These two metrics are then averaged over all samples and normalized by the sensitivity measured when both inputs are perturbed, yielding $d_{\mathrm{act}}(t)$ and $d_{\mathrm{ctx}}(t)$ as the final reliance scores on $x_t$ and $c$ at flow time $t$, respectively.

% \textcolor{red}{Appendix: Report the boundary-estimation protocol concisely: 256 chunks per run, partners sampled from different episodes, five independent runs or partner maps, and 24 flow-time values. Also specify $\sigma$, the classifier split, and whether the AUC curve is smoothed; otherwise the “first” chance-level crossing may be sensitive to sampling noise.}

\begin{figure}[t]
    \centering
        \vspace{-0.1in}
    \makebox[\linewidth][c]{%
        \includegraphics[width=1.0\linewidth]{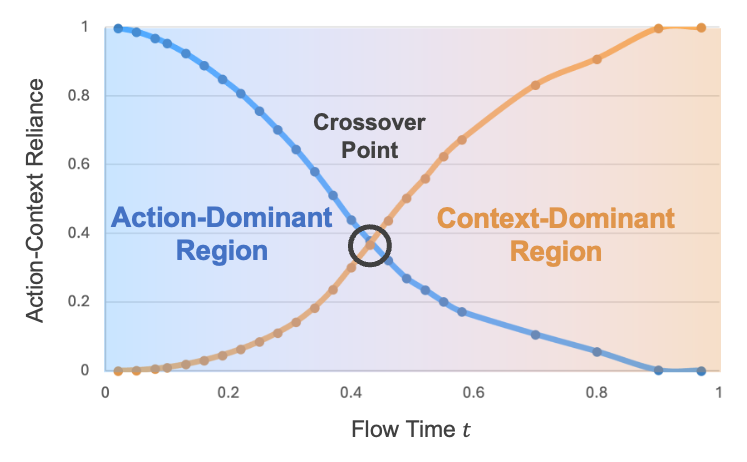}
    }
    \vspace{-0.4in}
% \caption{Interventional estimate of policy input reliance across flow time. Curves show normalized action-space sensitivity to replacing the noised action ($S^{x}=d_{\mathrm{act}}/D$) and task context ($S^{o}=d_{\mathrm{ctx}}/D$) on 256 action chunks from 100 episodes. The curves cross at $t^{\star}=0.4327$ (95\% bootstrap CI $0.4067,0.4519$); shaded bands denote pointwise 95\% CIs, and the dashed line and gray band mark the crossing and its CI. Across five fixed partner assignments, $t^{\star}\in[0.4327,0.4532]$. \textcolor{red}{refine caption and text}
\caption{Empirical analysis of action--context reliance. The blue curve represents action reliance, while the orange curve represents context reliance. For this setup, the two curves intersect at $t^{\star}=0.4327$.}

    \label{fig:action_context}
\end{figure}

 Figure~\ref{fig:action_context} shows a single crossover from action-dominant prediction at low flow times to context-dominant prediction at high flow times. Therefore, at low flow times, $\mathcal{D}_{\mathrm{tm}}$ can provide precise action supervision with limited influence from task mismatch; at high flow times, the target-aligned task context in $\mathcal{D}_{\mathrm{pm}}$ becomes more beneficial despite its lower action precision.

\subsection{Offline Admission Boundaries}
\label{sec:boundaries}

The observed reliance transition motivates admitting $\mathcal{D}_{\mathrm{tm}}$ at low flow times and $\mathcal{D}_{\mathrm{pm}}$ at high flow times. However, since the two sources differ from $\mathcal{D}_{\mathrm{pt}}$ along
different axes, their boundaries require different diagnostics: action separability for $\mathcal{D}_{\mathrm{pm}}$ and policy action-context reliance for $\mathcal{D}_{\mathrm{tm}}$. Therefore, we define:
\begin{equation}
\mathcal{T}_{\mathrm{tm}}=[0,t_{\mathrm{tm}}],
\qquad
\mathcal{T}_{\mathrm{pm}}=[t_{\mathrm{pm}},1]
\end{equation}
where $t_{\mathrm{tm}}$ and $t_{\mathrm{pm}}$ are the admission boundaries.

% The two imperfect sources require different boundary diagnostics since each differs from $\mathcal{D}_{\mathrm{pt}}$ along a different axis. $\mathcal{D}_{\mathrm{pm}}$ shares the task intent but has lower action precision, making it more suitable at high flow times, where task context is more influential and source-specific action differences are less visible in $x_t$. In contrast, $\mathcal{D}_{\mathrm{tm}}$ provides precise actions from different tasks, making it
% more suitable at low flow times, where action information dominates and task context has less influence. Accordingly, we use a data-level action-separability diagnostic for $\mathcal{D}_{\mathrm{pm}}$ and a policy-level action-context
% reliance diagnostic for $\mathcal{D}_{\mathrm{tm}}$. We parameterize their admission regions as:
% %
% \begin{equation}
% \mathcal{T}_{\mathrm{tm}}=[0,t_{\mathrm{tm}}],
% \qquad
% \mathcal{T}_{\mathrm{pm}}=[t_{\mathrm{pm}},1]
% \end{equation}
% %
% where $t_{\mathrm{tm}}$ and $t_{\mathrm{pm}}$ are the admission boundaries.

% We therefore train a flow-time-conditioned classifier to predict whether $x_t$ comes from $\mathcal{D}_{\mathrm{pm}}$ or $\mathcal{D}_{\mathrm{pt}}$, where $x_t$ is constructed by adding different level of noises to clean action.  as shown in Figure~\ref{fig:method} (left).

\textbf{Task-Matched, Precision-Mismatched Boundary.}
Since $\mathcal{D}_{\mathrm{pm}}$ and $\mathcal{D}_{\mathrm{pt}}$ demonstrate the same task, the separability of their noised actions primarily reflects differences in action precision. We therefore train a flow-time-conditioned source classifier to predict whether $x_t$ originates from $\mathcal{D}_{\mathrm{pm}}$ or $\mathcal{D}_{\mathrm{pt}}$. Here, $x_t$ is constructed by perturbing the corresponding clean action with a flow-time-dependent noise level following Equation~\ref{eq:ot-path}, as illustrated in Figure~\ref{fig:method} (left).

Following classifier-based two-sample testing~\cite{lopezpaz2017revisiting} and source-separability
diagnostics~\cite{xdiffusion}, we use its held-out AUC to define:
\begin{equation}
t_{\mathrm{pm}}
=
\min\left\{
t\in[0,1]:
\left|
\operatorname{AUC}(t)-0.5
\right|
\leq
\sigma
\right\}
\label{eq:tpm}
\end{equation}
where $\operatorname{AUC}=0.5$ indicates that the classifier cannot identify the data source better than random guessing, and $\sigma$ specifies the allowed deviation from this value. We fix $\sigma = 0.05$ across all tasks without tuning on downstream policy performance.
 Thus, $t_{\mathrm{pm}}$ is the earliest flow time at which the precision mismatch is no longer reliably detectable from $x_t$. We set $\mathcal{T}_{\mathrm{pm}}
=[t_{\mathrm{pm}},1]$, retaining its target-aligned task context in the high-flow-time region while reducing the influence of source-dependent action differences in the noised input.

\textbf{Precision-Matched, Task-Mismatched Boundary.}
The action-separability diagnostic used for $\mathcal{D}_{\mathrm{pm}}$ is not suitable for $\mathcal{D}_{\mathrm{tm}}$. Because different tasks naturally induce different action distributions, a source classifier may distinguish $
\mathcal{D}_{\mathrm{tm}}$ from $\mathcal{D}_{\mathrm{pt}}$ even when their action precision is matched. The distinction should be reflected at the task context level rather than at the action level. We reuse action-context reliance derived from Section~\ref{sec:input-reliance} to identify when task context becomes more influential than the noised action input.

At low flow times, prediction relies primarily on the noised action input, allowing $\mathcal{D}_{\mathrm{tm}}$ to provide precise action supervision with limited influence from task mismatch. As flow time increases, context becomes
more influential, making the task mismatch more consequential. Given the observed single crossover between the two reliance curves, we define $t_{\mathrm{tm}}$ as the latest flow time at which action reliance remains at least as strong
as context reliance as shown in Figure~\ref{fig:method} (right):
\begin{equation}
t_{\mathrm{tm}}
=
\max\left\{
t\in[0,1]:
d_{\mathrm{act}}(t)\geq d_{\mathrm{ctx}}(t)
\right\}
\label{eq:ttm}
\end{equation}
We then set $\mathcal{T}_{\mathrm{tm}}=[0,t_{\mathrm{tm}}]$, retaining $\mathcal{D}_{\mathrm{tm}}$ in the action-dominant regime while limiting the influence of its task-mismatched context.

% LEAVE THIS TO APPENDIX, THIS BREAKS THE FLOW. hao:ok .Unlike $t_{\mathrm{pm}}$, which is a property of the data sources,
% $t_{\mathrm{tm}}$ is a property of the target policy. We therefore evaluate the
% probe on the target checkpoint used for the reported experiments. As the policy
% converges during post-training, the probe crossing stabilizes, so this choice is
% not sensitive to the exact step (Appendix~\ref{app:probe}).

% In practice, we estimate $t_{\mathrm{tm}}$ by linearly interpolating between the two evaluated flow times surrounding the crossover.

\subsection{Policy Integration}
\label{sec:policy-integration}

Given the computed admission boundaries, $\boldsymbol{\varepsilon}\mathrm{4P}$ modifies policy training through source-dependent flow-time sampling. At each training step, we sample a source $s\sim\pi$ and an example $(a,c)\sim\mathcal{D}_s$, and then sample $t$
uniformly from the corresponding admission region:
\begin{equation}
\mathcal{T}_s =
\begin{cases}
[0,1], & s=\mathrm{pt}\\
[t_{\mathrm{pm}},1], & s=\mathrm{pm}\\
[0,t_{\mathrm{tm}}], & s=\mathrm{tm}
\end{cases}
\label{eq:policy-integration}
\end{equation}

The sampled $t$ is used to construct $x_t$ according to Equation~\eqref{eq:ot-path}, after which the policy is optimized with the standard flow-matching objective. Thus, $\mathcal{D}_{\mathrm{pt}}$ supervises the full flow trajectory,
while each imperfect source contributes only within its admitted region. The source sampling distribution $\pi$, model architecture, and training objective remain unchanged. The admission boundaries are computed once offline and remain fixed throughout training.

\section{Experimental Setup}

\textbf{Hardware Setup.}
\label{sec:hardware}
All experiments are conducted using the Rainbow Robotics RB-Y1 robot. High-precision demonstrations are collected through robot teleoperation, while lower-precision demonstrations are collected using our self-built UMI system. For both teleoperation and UMI data, the policy uses visual observations from two Intel RealSense D405 cameras. The UMI camera placement and gripper configuration are matched to those of the robot to minimize differences in viewpoint and embodiment.

\begin{figure}[t]
    \centering
    \makebox[\linewidth][c]{%
        \includegraphics[width=1.05\linewidth]{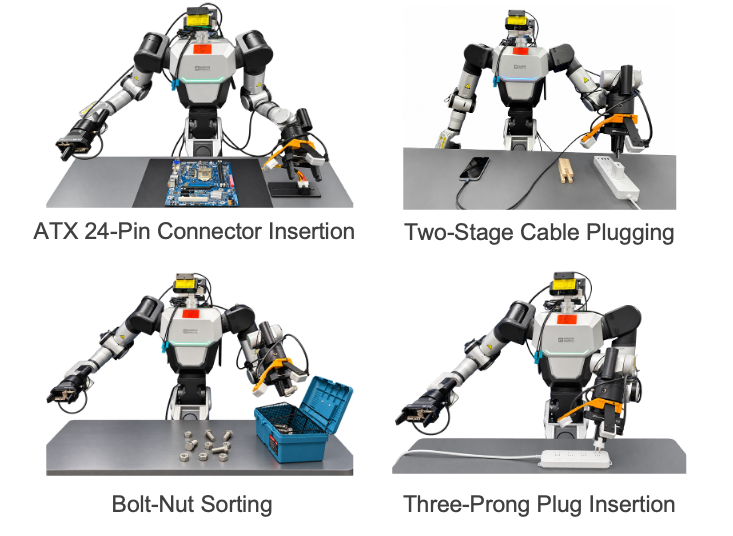}
    }
    \caption{Real-robot tasks used in the experiments. The target tasks are ATX 24-Pin Connector Insertion, Two-Stage Cable Plugging, and Bolt-Nut Sorting; Three-Prong Plug Insertion serves as the shared auxiliary dataset $\mathcal{D}_{\mathrm{tm}}$ across three target tasks.}

    \label{fig:task_overview}
\end{figure}

\textbf{Tasks and Data.}
All methods are evaluated on three real-robot manipulation tasks spanning different precision requirements: (1) ATX 24-Pin Connector Insertion requires inserting a 24-pin ATX connector into a motherboard header, with approximately 0.2~mm nominal clearance between each 3.3~mm pin and its 3.5~mm opening. (2) Two-Stage Cable Plugging requires first seating a two-prong plug and then inserting the USB-A end. The second stage requires approximately 0.3~mm nominal lateral clearance. (3) Bolt-Nut Sorting requires separating mixed bolts and nuts into their corresponding bins. This task has substantially larger placement tolerance and serves as a coarse-grained manipulation benchmark to assess whether the benefits of our method are specific to high-precision tasks. Together, these tasks evaluate performance across manipulation settings ranging from sub-millimeter-precision insertion to coarse object sorting. For brevity, we refer to the three tasks as \texttt{ATX}, \texttt{Cable}, and \texttt{Sorting}, respectively. Figure~\ref{fig:task_overview} gives an overview of the three target tasks.

  \begin{table}[t]
      \centering
      \caption{Composition of the real-robot training datasets. }
      {
      \small
      \setlength{\tabcolsep}{13 pt}
      \renewcommand{\arraystretch}{1.5}
      \begin{tabular}{@{}llccc@{}}
          \toprule
          \multirow{2}{*}{\textbf{Dataset}}
          & \multirow{2}{*}{\textbf{Type}}
          & \multicolumn{3}{c}{\textbf{Demonstrations}} \\
          \cmidrule(lr){3-5}
          & & \textbf{ATX} & \textbf{Sorting} & \textbf{Cable} \\
          \midrule
          $\mathcal{D}_{\mathrm{pt}}$
          & Teleop & 300 & 100 & 100 \\

          $\mathcal{D}_{\mathrm{pm}}$
          & UMI & 100 & 50 & 50 \\

          $\mathcal{D}_{\mathrm{tm}}$
          & Teleop & 100 & 50 & 50 \\

          $\mathcal{D}_{\mathrm{pt}}^{+}$
          & Teleop & 100 & 50 & 50 \\
          \bottomrule
      \end{tabular}
      }
      
      \label{tab:dataset_composition}
  \end{table}

% For task-mismatched high-precision data $\mathcal{D}_{\mathrm{tm}}$, we use demonstrations of inserting a black three-prong plug. 
% Post-training uses three datasets $\mathcal{D}_{\mathrm{pt}}$, $\mathcal{D}_{\mathrm{pm}}$,
% and $D_{\mathrm{tm}}$.
% We additionally collect an extra set of
% high-precision target-task robot demonstrations $\mathcal{D}_{\mathrm{ept}}$, used only
% as a reference baseline to compare $\boldsymbol{\varepsilon}\mathrm{4P}$ against collecting an equal
% amount of additional high-quality target-task teleoperation. 
% As shown in Table~\ref{tab:rq1_defects}, 
For precision-mismatched data, we use UMI to collect $\mathcal{D}_{\mathrm{pm}}$ for each target task. For task-mismatched data, we collect $\mathcal{D}_{\mathrm{tm}}$ of inserting a three-prong plug, and use it as a shared auxiliary dataset across all three target tasks. We additionally collect $\mathcal{D}_{\mathrm{pt}}^{+}$, a supplementary set of high-precision target-task demonstrations used to construct reference training sets for comparisons in which a portion of high-quality data is replaced with imperfect data. Table~\ref{tab:dataset_composition} summarizes the collection source and number of demonstrations for each dataset and target task.

\textbf{Evaluation.}
We adopt the pre-trained $\pi_{0.5}$~\cite{pi05,openpi} as the base policy, although $\boldsymbol{\varepsilon}\mathrm{4P}$ can be readily extended to other flow-matching policies. All evaluated methods use the same model architecture, pre-trained checkpoint, optimizer, batch size, source-sampling probabilities, and data and training budgets, and differ only in their source-dependent flow-time admission boundaries. Unless otherwise specified, we report the binary success rate over 60 real-robot trials per task.

% ============================================================
% RQ1
% ===========================================================

% ============================================================
% RQ3: Is the location of the window important?
% ============================================================

\newcommand{\result}[3]{%
    $#1_{\scriptscriptstyle #2}^{\scriptscriptstyle #3}$%
}

\newcommand{\bestresult}[3]{%
    $\mathbf{#1}_{\scriptscriptstyle #2}^{\scriptscriptstyle #3}$%
}

\section{Experiments}

We structure our experiments around four research questions to evaluate the effectiveness of the proposed method and understand how it leverages imperfect data beyond success rates:
  \begin{itemize}
  \item Under native co-training, does imperfect data help or harm policy learning?
  \item Can $\boldsymbol{\varepsilon}\mathrm{4P}$ effectively upcycle imperfect data to improve policy performance?
  \item Do the identified admission regions effectively capture where imperfect data is useful?
  \item Can $\boldsymbol{\varepsilon}\mathrm{4P}$ replace a portion of high-quality data with imperfect data?
  \end{itemize}

% ============================================================
% RQ1
% ============================================================
\subsection{Imperfect Data Is Harmful under Native Co-Training}
 \begin{table}[t]
      \centering
      \caption{
          Real-robot success rates (\%) under native co-training, where each data source is admitted over the full flow-time axis $t\in[0,1]$.
      }
      \small
      \setlength{\tabcolsep}{3pt}
      \renewcommand{\arraystretch}{1.5}

      \begin{tabular*}{\columnwidth}
          {@{\extracolsep{\fill}}lccc@{}}
          \toprule
          \textbf{Training data}
          & \textbf{ATX}
          & \textbf{Sorting}
          & \textbf{Cable} \\
          \midrule

          $\mathcal{D}_{\mathrm{pt}}$
          & \best{48.3}
          & \best{81.7}
          & \best{78.3} \\

          $\mathcal{D}_{\mathrm{pt}}+\mathcal{D}_{\mathrm{pm}}$
          & 33.3
          & 75.0
          & 25.0 \\

          $\mathcal{D}_{\mathrm{pt}}+\mathcal{D}_{\mathrm{tm}}$
          & 43.3
          & 78.3
          & 71.7 \\

          \bottomrule
      \end{tabular*}

      \label{tab:rq1_defects}
  \end{table}
  
Before evaluating the effectiveness of $\boldsymbol{\varepsilon}\mathrm{4P}$, we first empirically examine whether imperfect data provides useful or harmful supervision under native co-training, where each source is admitted over the full flow-time axis, i.e., $\mathcal{T}_s=[0,1]$ for all $s\in\{\mathrm{pt},\mathrm{pm},\mathrm{tm}\}$. As shown in Table~\ref{tab:rq1_defects}, training only on $\mathcal{D}_{\mathrm{pt}}$ establishes baseline success rates of 48.3\%, 81.7\%, and 78.3\% on \texttt{ATX}, \texttt{Sorting}, and \texttt{Cable}, respectively.
Native co-training with $\mathcal{D}_{\mathrm{pm}}$ substantially degrades performance on all three tasks, but the magnitude of degradation depends strongly on the precision requirement of the target task. Specifically, across the two high-precision tasks \texttt{ATX} and \texttt{Cable}, the success rate is reduced by 34.2 percentage points on average, compared with only a 6.7-percentage-point decrease on the coarse-grained \texttt{Sorting} task. This pattern suggests that lower-precision action supervision becomes increasingly harmful as successful task completion depends more strongly on fine-grained execution precision.

Co-training with $\mathcal{D}_{\mathrm{tm}}$ exhibits a different failure pattern. Although it also reduces performance across all three tasks, the degradation is substantially smaller: 5.0 percentage points on \texttt{ATX}, 3.4 points on \texttt{Sorting}, and 6.6 points on \texttt{Cable}. Thus, $\mathcal{D}_{\mathrm{tm}}$ introduces less severe degradation despite differing from the target tasks in context. This comparison suggests that different forms of mismatch affect native co-training in fundamentally different ways. Task-mismatched data behaves more like auxiliary multi-task supervision while preserving high-precision execution, whereas precision-mismatched data introduces imprecise action supervision directly on the target task, making it substantially more harmful for high-precision manipulation. 

% ============================================================
% RQ2
% ============================================================
\subsection{$\boldsymbol{\varepsilon}\mathrm{4P}$ Upcycles Imperfect Data to Improve Performance}

\begin{table}[t]
        \centering
        \caption{
            Real-robot success rates (\%) with native co-training and $\boldsymbol{\varepsilon}\mathrm{4P}$. Bold and underlined values denote the best and second-best results, respectively, within each column.
        }
        \footnotesize
        \setlength{\tabcolsep}{3pt}
        \renewcommand{\arraystretch}{1.5}

        \begin{tabular}{@{}llcccc@{}}
            \toprule
            \multirow{2}{*}{\textbf{Training data}}
            & \multirow{2}{*}{\textbf{Admission}}
            & \multirow{2}{*}{\textbf{ATX}}
            & \multirow{2}{*}{\textbf{Sorting}}
            & \multicolumn{2}{c}{\textbf{Cable}} \\
            \cmidrule(lr){5-6}
            & & & & \textbf{Stage 1} & \textbf{Stages 1--2} \\
            \midrule

            $\mathcal{D}_{\mathrm{pt}}$
            & $\mathcal{T}_{\mathrm{pt}}$
            & 48.3 & 81.7 & 90.0 & 78.3 \\

            \midrule

            $\mathcal{D}_{\mathrm{pt}}+\mathcal{D}_{\mathrm{pm}}$
            & $[0,1]$
            & 33.3 & 75.0 & 70.0 & 25.0 \\

            $\mathcal{D}_{\mathrm{pt}}+\mathcal{D}_{\mathrm{tm}}$
            & $[0,1]$
            & 43.3 & 78.3 & 86.7 & 71.7 \\

            $\mathcal{D}_{\mathrm{pt}}+\mathcal{D}_{\mathrm{pm}}
            +\mathcal{D}_{\mathrm{tm}}$
            & $[0,1]$
            & 38.3 & 73.3 & 75.0 & 33.3 \\

            \midrule

            $\mathcal{D}_{\mathrm{pt}}+\mathcal{D}_{\mathrm{pm}}$
            & $\mathcal{T}_{\mathrm{pt}},\mathcal{T}_{\mathrm{pm}}$
            & \underline{73.3}
            & \textbf{93.3}
            & \underline{98.3}
            & \textbf{90.0} \\

            $\mathcal{D}_{\mathrm{pt}}+\mathcal{D}_{\mathrm{tm}}$
            & $\mathcal{T}_{\mathrm{pt}},\mathcal{T}_{\mathrm{tm}}$
            & 71.7
            & 88.3
            & \textbf{100.0}
            & 81.7 \\

            $\mathcal{D}_{\mathrm{pt}}+\mathcal{D}_{\mathrm{pm}}
            +\mathcal{D}_{\mathrm{tm}}$
            & $\mathcal{T}_{\mathrm{pt}},\mathcal{T}_{\mathrm{pm}},
            \mathcal{T}_{\mathrm{tm}}$
            & \textbf{80.0}
            & \underline{91.7}
            & \underline{98.3}
            & \underline{88.3} \\

            \bottomrule
        \end{tabular}

        \label{tab:main_results}
    \end{table}

Having shown that admitting imperfect data over the full flow-time axis causes source-dependent performance degradation, we next evaluate whether $\boldsymbol{\varepsilon}\mathrm{4P}$ can turn this otherwise harmful supervision into performance gains by restricting each source to its identified admission region. Table~\ref{tab:main_results} summarizes the results.

% pm & tm
For $\mathcal{D}_{\mathrm{pm}}$, restricting admission to $\mathcal{T}_{\mathrm{pm}}$ increases success on \texttt{ATX} from 33.3\% to 73.3\%, with substantial gains also observed on \texttt{Sorting} and the full \texttt{Cable} task. Similarly, restricting $\mathcal{D}_{\mathrm{tm}}$ to $\mathcal{T}_{\mathrm{tm}}$ increases success on \texttt{ATX} from 43.3\% to 71.7\% and consistently improves performance on the other tasks. The gain is particularly pronounced on the more challenging \texttt{ATX} task. Interestingly, $\mathcal{D}_{\mathrm{pm}}$ is more harmful than $\mathcal{D}_{\mathrm{tm}}$ under native co-training, yet performs better across all three tasks under
$\boldsymbol{\varepsilon}\mathrm{4P}$. This reversal suggests that $\mathcal{D}_{\mathrm{pm}}$ is highly useful but more sensitive to where it is admitted during training.
  
% tm + pm
We then admit both imperfect sources within their respective admission regions. The resulting policy achieves 80.0\% success on \texttt{ATX}, 91.7\% on \texttt{Sorting}, and 88.3\% full-task success on \texttt{Cable}, consistently outperforming both native co-training with the same sources and training with $\mathcal{D}_{\mathrm{pt}}$ alone. The benefit of composition is clearest on \texttt{ATX}, where it outperforms either imperfect source alone. On \texttt{Sorting} and \texttt{Cable}, the best single-source policies already approach the performance ceiling, and composition therefore does not further improve their success rates. Nevertheless, it retains nearly all of their gains, suggesting that the two heterogeneous imperfect sources can occupy complementary flow-time regions within a single training run. 

\begin{figure*}[t]
    \centering
    \includegraphics[width=\textwidth]{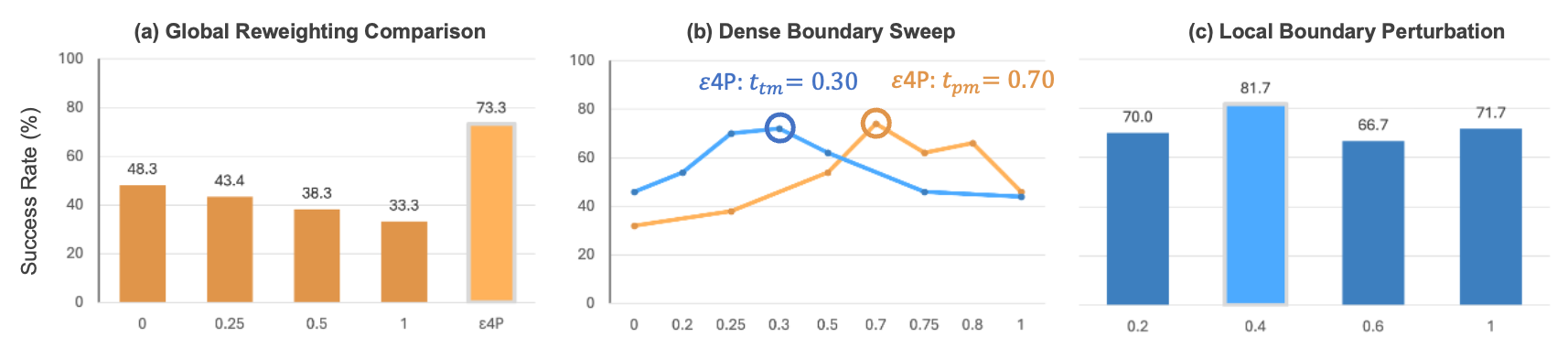}
    \caption{(a) Global loss reweighting comparison using $\mathcal{D}_{\mathrm{pm}}$ on \texttt{ATX}. (b) Dense boundary sweep for $t_{\mathrm{pm}}$ and $t_{\mathrm{tm}}$ on \texttt{ATX}. (c) Local boundary perturbation of $t_{\mathrm{tm}}$ on \texttt{Cable}.}

    \label{fig:sweep}
\end{figure*}

% reweighting
As an additional global loss reweighting baseline, we test whether simply reducing the influence of imperfect data can match the gains of source-dependent admission. On \texttt{ATX}, we fix the weight of $\mathcal{D}_{\mathrm{pt}}$ to $w_{\mathrm{pt}}=1$ and keep $\mathcal{D}_{\mathrm{pm}}$ active over the full flow-time range $[0,1]$, while varying $w_{\mathrm{pm}}\in\{0,0.25,0.5,1\}$. Here, $w_{\mathrm{pm}}=0$ corresponds to using $\mathcal{D}_{\mathrm{pt}}$ alone, whereas $w_{\mathrm{pm}}=1$ corresponds to native co-training. As shown in Figure~\ref{fig:sweep}(a), reducing $w_{\mathrm{pm}}$ mitigates the degradation caused by native co-training, but no nonzero weight outperforms the $\mathcal{D}_{\mathrm{pt}}$-only baseline. In contrast, $\boldsymbol{\varepsilon}\mathrm{4P}$ retains $w_{\mathrm{pm}}=1$ and restricts only its flow-time support, achieving the highest 73.3\% success. This comparison shows that the benefit comes from controlling where, rather than merely how much, imperfect data contributes.

% ============================================================
% RQ3
% ============================================================
\subsection{$\boldsymbol{\varepsilon}\mathrm{4P}$ Identifies Effective Admission Regions}

 \begin{table}[t]
      \centering
      \caption{
         Real-robot success rates (\%) on \texttt{ATX} under three admission strategies: native co-training, opposite-window control, and $\boldsymbol{\varepsilon}\mathrm{4P}$. Bold values denote the best result for each source.
      }
      \footnotesize
      \setlength{\tabcolsep}{8pt}
      \renewcommand{\arraystretch}{1.5}

      \begin{tabular}{@{}lllc@{}}
          \toprule
          \textbf{Imperfect data}
          & \textbf{Method}
          & \textbf{Admission}
          & \textbf{Success} \\
          \midrule

          \multirow{3}{*}{$\mathcal{D}_{\mathrm{pm}}$}
          & Native co-training
          & $[0,1]$
          & 33.3 \\

          & Opposite window
          & $[0,t_{\mathrm{pm}}]$
          & 30.0 \\

          & \ours{}
          & $[t_{\mathrm{pm}},1]$
          & \textbf{73.3} \\

          \midrule

          \multirow{3}{*}{$\mathcal{D}_{\mathrm{tm}}$}
          & Native co-training
          & $[0,1]$
          & 43.3 \\

          & Opposite window
          & $[t_{\mathrm{tm}},1]$
          & 46.7 \\

          & \ours{}
          & $[0,t_{\mathrm{tm}}]$
          & \textbf{71.7} \\

          \bottomrule
      \end{tabular}

      \label{tab:rq3_controls}
  \end{table}

We next examine whether the admission regions identified by $\boldsymbol{\varepsilon}\mathrm{4P}$ correspond to regions that are effective for policy learning. We evaluate this through three complementary experiments: (1) a dense boundary sweep to assess global agreement, (2) a local perturbation of the identified $t_{\mathrm{tm}}$ to assess whether this boundary is locally preferred, and (3) opposite-window controls to test whether the identified regions reflect meaningful source-dependent structure rather than arbitrary design choices. All boundaries are computed using the offline diagnostics; the downstream evaluations are used only for validation.

First, we assess whether the admission boundaries identified by $\boldsymbol{\varepsilon}\mathrm{4P}$ align with those favored by downstream policy performance. On the most challenging \texttt{ATX} task, we compare the identified $t_{\mathrm{pm}}$ and $t_{\mathrm{tm}}$ against the performance landscape obtained from dense boundary sweeps. As shown in Figure~\ref{fig:sweep}(b), both boundaries lie within the high-performing regions. This agreement supports the effectiveness of the admission regions identified by $\boldsymbol{\varepsilon}\mathrm{4P}$.

Second, we assess whether an identified boundary $t_{\mathrm{tm}}$ is also locally preferred on a different task. Rather than performing another dense sweep, we evaluate the identified $t_{\mathrm{tm}}=0.4$ on the \texttt{Cable} task together with two neighboring values, $t_{\mathrm{tm}}=0.2$ and $t_{\mathrm{tm}}=0.6$. As shown in Figure~\ref{fig:sweep}(c), the identified boundary achieves the highest success among the evaluated values, providing additional local validation on a second task.

Finally, we test whether the selected admission regions are meaningful or whether any truncation of the flow-time range would provide similar benefits. As shown in Table~\ref{tab:rq3_controls}, on \texttt{ATX}, $\mathcal{T}_{\mathrm{pm}}=[t_{\mathrm{pm}},1]$ achieves 73.3\% success, compared with 30.0\% using the opposite window $[0,t_{\mathrm{pm}}]$ and 33.3\% under native co-training. Similarly, $\mathcal{T}_{\mathrm{tm}}=[0,t_{\mathrm{tm}}]$ achieves 71.7\%, compared with 46.7\% using the opposite window  $[t_{\mathrm{tm}},1]$ and 43.3\% under native co-training. These results show that performance depends on where each source is admitted, not merely on truncating its flow-time region.

Together, the global sweep, local perturbation, and opposite-window controls show that $\boldsymbol{\varepsilon}\mathrm{4P}$ identifies effective source-dependent admission regions that closely align with those favored by downstream policy performance.

% ============================================================
% RQ4
% ============================================================

\subsection{$\boldsymbol{\varepsilon}\mathrm{4P}$ Reduces Reliance on High-Quality Data}

Lastly, we examine whether $\boldsymbol{\varepsilon}\mathrm{4P}$ can reduce reliance on high-quality data by comparing source-dependent admission of imperfect data with training on additional perfect data $\mathcal{D}_{\mathrm{pt}}^{+}$. For a controlled comparison, $\mathcal{D}_{\mathrm{pt}}^{+}$, $\mathcal{D}_{\mathrm{pm}}$, and $\mathcal{D}_{\mathrm{tm}}$ contain equal amounts of data. 

As shown in Table~\ref{tab:additional_perfect_data}, replacing $\mathcal{D}_{\mathrm{pt}}^{+}$ with $\mathcal{D}_{\mathrm{pm}}$ yields closely comparable performance. Source-dependent admission of $\mathcal{D}_{\mathrm{pm}}$ therefore recovers nearly all the benefit of augmenting $\mathcal{D}_{\mathrm{pt}}$ with an equal amount of perfect data. The results with $\mathcal{D}_{\mathrm{tm}}$ are less consistent, with gaps of 3.3 percentage points on \texttt{ATX}, 6.7 points on \texttt{Sorting}, and 10.0 points on the full \texttt{Cable} task. Thus, the extent to which imperfect data can replace high-quality data depends on the type and degree of mismatch. Overall, across both imperfect-data sources and all three tasks, $\boldsymbol{\varepsilon}\mathrm{4P}$ replaces an equal amount of additional high-quality data with suitably admitted imperfect data while incurring an average success-rate drop of only 4.2 percentage points.

 \begin{table}[t]
      \centering
      \caption{
Real-robot success rates (\%) comparing $\boldsymbol{\varepsilon}\mathrm{4P}$ with training on $\mathcal{D}_{\mathrm{pt}}^{+}$. Bold and underlined values denote the best and second-best results, respectively, within each column.
      }
      \footnotesize
      \setlength{\tabcolsep}{5pt}
      \renewcommand{\arraystretch}{1.5}

      \begin{tabular}{@{}llcccc@{}}
          \toprule
          \multirow{2}{*}{\textbf{Training data}}
          & \multirow{2}{*}{\textbf{Admission}}
          & \multirow{2}{*}{\textbf{ATX}}
          & \multirow{2}{*}{\textbf{Sorting}}
          & \multicolumn{2}{c}{\textbf{Cable}} \\
          \cmidrule(lr){5-6}
          & & & & \textbf{Stage 1} & \textbf{Stages 1--2} \\
          \midrule

          $\mathcal{D}_{\mathrm{pt}}$
          & $\mathcal{T}_{\mathrm{pt}}$
          & 48.3 & 81.7 & 90.0 & 78.3 \\

          $\mathcal{D}_{\mathrm{pt}}+\mathcal{D}_{\mathrm{pt}}^{+}$
          & $\mathcal{T}_{\mathrm{pt}}$
          & \textbf{75.0}
          & \textbf{95.0}
          & \textbf{100.0}
          & \textbf{91.7} \\

          $\mathcal{D}_{\mathrm{pt}}+\mathcal{D}_{\mathrm{pm}}$
          & $\mathcal{T}_{\mathrm{pt}},\mathcal{T}_{\mathrm{pm}}$
          & \underline{73.3}
          & \underline{93.3}
          & \underline{98.3}
          & \underline{90.0} \\

          $\mathcal{D}_{\mathrm{pt}}+\mathcal{D}_{\mathrm{tm}}$
          & $\mathcal{T}_{\mathrm{pt}},\mathcal{T}_{\mathrm{tm}}$
          & 71.7
          & 88.3
          & \textbf{100.0}
          & 81.7 \\

          \bottomrule
      \end{tabular}

      \label{tab:additional_perfect_data}
  \end{table}

\section{Discussion and Future Work}

\textbf{Discussion.} Our results suggest that imperfect auxiliary data should not be accepted
or rejected globally; its utility depends on where it contributes along
the flow-time axis. $\mathcal{D}_{\mathrm{pm}}$ and
$\mathcal{D}_{\mathrm{tm}}$ provide useful and harmful supervision at
different flow times, with their effective admission regions lying at
opposite ends of the axis. The opposite-window and loss reweighting
controls further show that the gains depend on \emph{where}, rather than
simply \emph{how much}, auxiliary supervision is applied.

The two admission boundaries capture distinct properties. The lower
boundary $t_{\mathrm{pm}}$ is primarily data-dependent and identifies
when source identity can no longer be reliably inferred from the noised
action input. In contrast, the upper boundary $t_{\mathrm{tm}}$ is
policy-dependent and reflects the relative sensitivity of the post-trained
policy to its action and conditioning inputs. Neither boundary should be
interpreted as an exact task-optimal threshold. Instead, they provide
practical offline criteria for locating effective admission regions
without exhaustive evaluation. The joint-training results further
suggest that heterogeneous sources can coexist within one policy when
assigned source-dependent admission regions.

\textbf{Limitations and Future Work.} Our experiments are currently limited to a single robot platform and a relatively small number of demonstrations collected in a controlled multi-camera setup. Although the experiments cover both high-precision and coarse-grained manipulation tasks, they do not yet fully evaluate the scalability and cross-embodiment generality of $\boldsymbol{\varepsilon}\mathrm{4P}$. Future work will evaluate $\boldsymbol{\varepsilon}\mathrm{4P}$ across a broader range of robot platforms and hardware configurations, and train policies with substantially larger and more diverse robot and human demonstration datasets.

% The current $t_{\mathrm{pm}}$ criterion measures compatibility only in
% the classifier's input space. Under the flow path in
% Equation~\ref{eq:ot-path}, the target velocity is
% $v^\star=\epsilon-a$. Consequently, the gradient induced by an admitted
% sample is
% %
% \begin{equation}
% g_\theta(a,c,\epsilon,t)
% =
% \nabla_\theta
% \left\|
% \hat v_\theta(x_t,c,t)-(\epsilon-a)
% \right\|_2^2 
% \label{eq:sample-gradient}
% \end{equation}
% %
% The demonstrated action $a$ therefore remains part of the training
% target at every flow time for which the sample is admitted. Even when
% actions from $\mathcal{D}_{\mathrm{pm}}$ and
% $\mathcal{D}_{\mathrm{pt}}$ become difficult to distinguish, they may
% still produce conflicting training gradients. Thus,
% $t_{\mathrm{pm}}$ should be viewed as an effective empirical proxy
% rather than a guarantee of gradient-level compatibility. A promising direction is to estimate $t_{\mathrm{pm}}$ directly from
% flow-time-dependent alignment or conflict between auxiliary and target
% gradients. Both boundaries could also be updated online during training: $t_{\mathrm{tm}}$ could be periodically recomputed as the
% policy evolves, while a future gradient-based $t_{\mathrm{pm}}$ could
% be re-estimated using the current policy. This would make
% $\boldsymbol{\varepsilon}\mathrm{4P}$ an adaptive rather than static
% admission mechanism.

\section{Conclusion}
We present $\boldsymbol{\varepsilon}\mathrm{4P}$, a source-dependent flow-time admission method for upcycling imperfect data in high-precision manipulation. Rather than uniformly co-training or globally reweighting imperfect sources, $\boldsymbol{\varepsilon}\mathrm{4P}$ admits precision-mismatched data at high flow times and task-mismatched
data at low flow times, using boundaries derived from offline diagnostics. Real-robot experiments across three tasks show that this strategy turns otherwise harmful supervision into performance gains, improving success rates by up to 31.7 percentage points over training with perfect target-task data alone. Moreover, across both imperfect-data sources and all three tasks, $\boldsymbol{\varepsilon}\mathrm{4P}$ replaces an equal amount of additional perfect data with an average performance drop of only 4.2 percentage points. These findings establish flow time as a useful axis for integrating heterogeneous imperfect data, pointing toward a scalable paradigm for high-precision manipulation.

\bibliographystyle{IEEEtran}
\bibliography{root}

\balance

% \clearpage
% \appendix
% \input{appendix}

\end{document}